\documentclass[a4paper]{article}

\usepackage[utf8]{inputenc}
\usepackage{subfigure}
\usepackage{adjustbox}
\usepackage{amssymb}
\usepackage{amsmath}
\DeclareMathOperator*{\argmax}{arg\,max}
\usepackage[algoruled]{algorithm2e}

\begin{document}
\title{Random Forest for Dissimilarity-based Multi-view Learning}
\author{
	Simon Bernard$^1$,
	Hongliu Cao$^{1,2}$, \\
	Robert Sabourin$^2$,
	Laurent Heutte$^1$ \\[1.4ex]
	$^1$ LITIS, Université de Rouen Normandie, 76000 Rouen, France\\[1ex]
	$^2$ LIVIA, École de Technologie Supérieure (ÉTS), \\
	Université du Québec, Montreal, QC, Canada \\[1.2ex]
	simon.bernard@univ-rouen.fr, caohongliu@gmail.com, \\ 
	robert.sabourin@etsmtl.ca, laurent.heutte@univ-rouen.fr
}
\date{}
\maketitle

\begin{abstract}
Many classification problems are naturally multi-view in the sense their data are described through multiple heterogeneous descriptions. For such tasks, dissimilarity strategies are effective ways to make the different descriptions comparable and to easily merge them, by (i) building intermediate dissimilarity representations for each view and (ii) fusing these representations by averaging the dissimilarities over the views. In this work, we show that the Random Forest proximity measure can be used to build the dissimilarity representations, since this measure reflects similarities between features but also class membership. We then propose a Dynamic View Selection method to better combine the view-specific dissimilarity representations. This allows to take a decision, on each instance to predict, with only the most relevant views for that instance. Experiments are conducted on several real-world multi-view datasets, and show that the Dynamic View Selection offers a significant improvement in performance compared to the simple average combination and two state-of-the-art static view combinations.
\end{abstract}

\section{Introduction}

In many real-world pattern recognition problems, the available data are complex in that they cannot be described by a single numerical representation. This may be due to multiple sources of information, as for autonomous vehicles for example, where multiple sensors are jointly used to recognize the environment \cite{Chen2017}. It may also be due to the use of several feature extractors, such as in image recognition tasks, often based on multiple representations of features, such as color, shape, texture descriptors, etc. \cite{Cao2019a} 

Learning from these types of data is called \textit{multi-view learning} and each modality/set of features is called a \textit{view}. For this type of task, it is assumed that the views convey different types of information, each of which can contribute to the pattern recognition task. Therefore, the challenge is generally to carry out the learning task taking into account the complementarity of the views. However, the difficulty with this is that these views can be very different from each other in terms of dimension, nature and meaning, and therefore very difficult to compare or merge. In a recent work \cite{Cao2019a}, we proposed to use dissimilarity strategies to overcome this issue. The idea is to use a dissimilarity measure to build intermediate representations from each view separately, and to merge them afterward. By describing the instances with their dissimilarities to other instances, the merging step becomes straightforward since the intermediate dissimilarity representations are fully comparable from one view to another.

For using dissimilarities in multi-view learning, two questions must be addressed: (i) how to measure and exploit the dissimilarity between instances for building the intermediate representation? and (ii) how to combine the view-specific dissimilarity representations for the final prediction? 

In our preliminary work \cite{Cao2019a}, the first question has been addressed with Random Forest (RF) classifiers. RF are known to be versatile and accurate classifiers \cite{Breim2001,Delga2014} but they are also known to embed a (dis)similarity measure between instances \cite{Englu2012}. The advantage of such a mechanism in comparison to traditional similarity measures is that it takes the classification/regression task into account for computing the similarities. For classification for example, the instances that belong to the same class are more likely to be similar according to this measure. Therefore, a RF trained on a view can be used to measure the dissimilarities between instances according to the view, and according to their class membership as well. The way this measure is used to build the per-view intermediate representations is by calculating the dissimilarity of a given instance $x$ to all the $n$ training instances. By doing so, $x$ can be represented by a new feature vector of size $n$, or in other words in a $n$-dimensional space where each dimension is the dissimilarity to one of the training instances. This space is called the dissimilarity space \cite{Pekal2005a, Costa2019} and is used as the intermediate representation for each view.

As for the second question, we addressed the combination of the view-specific dissimilarity representations by computing the average dissimilarities over all the views. That is to say, for an instance $x$, all the view-specific dissimilarity vectors are computed and averaged to obtain a final vector of size $n$. Each value in this vector is thus the average dissimilarity between $x$ and one of the $n$ training instances over the views. This is a simple, yet meaningful way to combine the information conveyed by each view. However, one could find it a little naive when considering the true rationale behind multi-view learning. Indeed, even if the views are expected to be complementary to each other, they are likely to contribute in very different ways to the final decision. One view in particular is likely to be less informative than another, and this contribution is even likely to be very different from an instance to predict to another. In that case, it is desirable to estimate and take this importance into account when merging the view-specific representations. This is the goal we follow in the present work.

In a nutshell, our preliminary work \cite{Cao2019a} has validated the generic framework explained above, with the two following key steps: (i) building the dissimilarity space with the RF dissimilarity mechanism and (ii) combining the views afterward by averaging the dissimilarities. In the present work, we deepen the second step by investigating two methods to better combine the view-specific dissimilarities: 
\begin{enumerate}
    \item combining the view-specific dissimilarities with a static weighted average, so that the views contribute differently to the final dissimilarity representation; in particular, we propose an original weight calculation method based on an analysis of the RF classifiers used to compute the view-specific dissimilarities;
    \item combining the view-specific dissimilarities with a dynamic combination, for which the views are solicited differently from one instance to predict to another; this dynamic combination is based on the definition of a \textit{region of competence} for which the performance of the RF classifiers is assessed and used for a view selection step afterward.
\end{enumerate}

The rest of this chapter is organized as follows. The Random Forest dissimilarity measure is firstly explained in Section \ref{sec:rfd}. The way it is used for multi-view classification is detailed in Section \ref{sec:dissim_mvl}. The different strategies for combining the dissimilarity representations are given in Section \ref{sec:combi}, along with our two proposals for static and dynamic view combinations. Finally, the experimental validation is presented in Section \ref{sec:xp}.

\section{Random Forest Dissimilarity}
\label{sec:rfd}

To fully understand the way a RF classifier can be used to compute dissimilarities between instances, it is first necessary to understand how an RF is built and how it gives a prediction for each new instance.

\subsection{Random Forest}
\label{ssec:rf}

In this work, the name "Random Forest" refers to the Breiman's reference method \cite{Breim2001}. Let us briefly recall its procedure to build a forest of $M$ Decision Trees, from a training set $T$. First, a bootstrap sample is built by random draw with replacement of $n$ instances, amongst the $n$ training instances available in $T$. Each of these bootstrap samples is then used to build one tree. During this induction phase, at each node of the tree, a splitting rule is designed by selecting a feature over $mtry$ features randomly drawn from the $m$ available features. The feature retained for the splitting rule at a given node is the one among the $mtry$ that maximizes the splitting criterion. At last, the trees in RF classifiers are grown to their maximum depth, that is to say when all their terminal nodes (also called leaves) are pure. The resulting RF classifier is typically noted as:
\begin{equation}
    H(\mathbf{x}) = \{ h_k(\mathbf{x}), k=1,\dots,M \}
\end{equation}
where $h_k(\mathbf{x})$ is the $k^{th}$ Random Tree of the forest, built using the mechanisms explained above \cite{Breim2001, Biau2016}. Note however that there exist many other RF learning methods that differ from the Breiman's method by the use of different randomization techniques for growing the trees \cite{Rokac2016}. \\

For predicting the class of a given instance $\mathbf{x}$ with a Random Tree, $\mathbf{x}$ goes down the tree structure from its root to one of its leaves. The descending path followed by $\mathbf{x}$ is determined by successive tests on the values of its features, one per node along the path. The prediction is given by the leaf in which $\mathbf{x}$ has landed. More information about this procedure can be found in the recently published RF reviews\cite{Verik2011, Biau2016, Rokac2016}. The key point here is that, if two test instances land in the same terminal node, they are likely to belong to the same class and they are also likely to share similarities in their feature vectors, since they have followed the same descending path. This is the main motivation behind using RF for measuring dissimilarities between instances.

Note that the final prediction of a RF classifier is usually obtained via majority voting over the component trees. Here again, there exist alternatives to majority voting \cite{Rokac2016}, but this latter remains the most used as far as we know.

\subsection{Using Random Forest for measuring dissimilarities}
\label{ssec:rfd}

The RF dissimilarity measure is the opposite measure of the RF proximity (or similarity) measure defined in Breiman's work \cite{Breim2001,Verik2011,Cao2019a}, the latter being noted $p_H(\mathbf{x}_i, \mathbf{x}_j)$ in the following. 

The RF dissimilarity measure is inferred from a RF classifier $H$, learned from $T$. Let us firstly define the dissimilarity measure inferred by a single Random Tree $h_k$, noted $d_k$: let $\mathcal{L}_k$ denote the set of leaves of $h_k$, and let $l_k(\mathbf{x})$ denote a function from the input domain $\mathcal{X}$ to $\mathcal{L}_k$, that returns the leaf of $h_k$ where $\mathbf{x}$ lands when one wants to predict its class. The dissimilarity measure $d_k$ is defined as in Equation \ref{eq:dissDT}: if two training instances $\mathbf{x}_i$ and $\mathbf{x}_j$ land in the same leaf of $h_k$, then the dissimilarity between both instances is set to $0$, else it is equal to $1$.
\begin{equation}
    \label{eq:dissDT}
    d_k(\mathbf{x}_i, \mathbf{x}_j) = 
    \left\{ 
        \begin{array}{ll}
            0, & if \  l_k(\mathbf{x}_i)=l_k(\mathbf{x}_j) \\
            1, & otherwise
        \end{array}
    \right.
\end{equation}
The $d_k$ measure is the strict opposite of the tree proximity measure $p_k$ \cite{Breim2001, Verik2011}, i.e. $d_k(\mathbf{x}_i, \mathbf{x}_j) = 1 - p_k(\mathbf{x}_i, \mathbf{x}_j)$.

Now, the measure $d_H(\mathbf{x}_i, \mathbf{x}_j)$ derived from the whole forest consists in calculating $d_k$ for each tree in the forest, and in averaging the resulting dissimilarity values over the $M$ trees, as follows:
\begin{equation}
    \label{eq:dissRF}
    d_H(\mathbf{x}_i, \mathbf{x}_j) = \frac{1}{M} \sum_{k=1}^{M} d_k(\mathbf{x}_i, \mathbf{x}_j)
\end{equation}
Similarly to the way the predictions are given by a forest, the rationale is that the accuracy of the dissimilarity measure $d_H$ relies essentially on the averaging over a large number of trees. Moreover, this measure is a pairwise function $d_H: \mathcal{X} \times \mathcal{X} \rightarrow \mathbb{R^+}$ that satisfies the reflexivity property ($d_H(\mathbf{x}_i, \mathbf{x}_i) = 0$), the non-negativity property ($d_H(\mathbf{x}_i, \mathbf{x}_j) \geq 0$) and the symmetry property ($d_H(\mathbf{x}_i, \mathbf{x}_j) = d_H(\mathbf{x}_j, \mathbf{x}_i)$). Note however that it does not satisfy the last two properties of the distance functions, namely the definiteness property ($d_H(\mathbf{x}_i, \mathbf{x}_j) = 0$ does not imply $\mathbf{x}_i = \mathbf{x}_j$) and the triangle inequality ($d_H(\mathbf{x}_i, \mathbf{x}_k)$ is not necessarily less or equal to $d_H(\mathbf{x}_i, \mathbf{x}_j) + d_H(\mathbf{x}_j, \mathbf{x}_k)$).

As far as we know, only few variants of this measure have been proposed in the literature \cite{Englu2012,Cao2019b}. These variants differ from the measure explained above in the way they infer the dissimilarity value from a tree structure. The motivation is to design a finer way to measure the dissimilarity than the coarse binary value given in Equation \ref{eq:dissDT}. This coarse value may seem intuitively too superficial to measure dissimilarities, especially considering that a tree structure can provide richer information about the way two instances are similar to each other. 

The first variant \cite{Englu2012} modifies the $p_H$ measure by using the path length from one leaf to another when two instances land in different leaf nodes. In this way, $p_k(\mathbf{x}_i, \mathbf{x}_j)$ does not take its value in $\{0,1\}$ anymore but is computed as follows:
\begin{equation}
    \label{eq:englund}
    p_H(\mathbf{x}_i, \mathbf{x}_j) = \frac{1}{M} \sum_{k=1}^{M} p_k(\mathbf{x}_i, \mathbf{x}_j) = \frac{1}{M} \sum_{k=1}^{M} \frac{1}{\exp{(w.g_{ijk})}}
\end{equation}
where, $g_{ijk}$ is the number of tree branches between the two terminal nodes occupied by $\mathbf{x}_i$ and $\mathbf{x}_j$ in the $k^{th}$ tree of the forest, and where $w$ is a parameter to control the influence of $g$ in the computation. When $l_k(\mathbf{x}_i)=l_k(\mathbf{x}_j)$, $d_k(\mathbf{x}_i, \mathbf{x}_j)$ is still equal to $0$, but in the opposite situation the resulting value is in $]0,1]$.

A second variant \cite{Cao2019b}, noted RFD in the following, leans on a measure of \textit{instance hardness}, namely the $\kappa$-Disagreeing Neighbors ($\kappa$DN) measure \cite{Smith2014}, that estimates the intrinsic difficulty to predict an instance as follows:
\begin{equation}
    \label{eq:kdn}
    \kappa DN(\mathbf{x}_i) = \frac{|\mathbf{x}_k: \mathbf{x}_k \in \kappa NN(\mathbf{x}_i) \cap y_k \neq y_i|}{\kappa}
\end{equation}
where $\kappa NN(\mathbf{x}_i)$ is the set of the $\kappa$ nearest neighbors of $\mathbf{x}_i$. This value is used for measuring the dissimilarity $\hat{d}_k(\mathbf{x},\mathbf{x}_i)$, between any instance $\mathbf{x}$ to any of the training instances $\mathbf{x}_i$, as follows:
\begin{equation}
    \label{eq:rfdih}
    \hat{d}_k(\mathbf{x},\mathbf{x}_i)=\frac{\sum_{k=1}^{M}(1-\kappa DN_k(\mathbf{x}_i)) \times d_k(\mathbf{x},\mathbf{x}_i)}{\sum_{k=1}^{M}(1-\kappa DN_k(\mathbf{x}_i))}
\end{equation}
where $\kappa DN_k(\mathbf{x}_i))$ is the $\kappa DN$ measure computed in the subspace formed by the sole features used in the $k^{th}$ tree of the forest.

Any of these variants could be used to compute the dissimilarities in our framework. However, we choose to use the RFD variant in the following, since it has been shown to give very good results when used for building dissimilarity representations for multi-view learning \cite{Cao2019b}.

\section{The dissimilarity representation for multi-view learning}
\label{sec:dissim_mvl}

\subsection{The dissimilarity space}
\label{ssec:dissim_space}

Among the different dissimilarity strategies for classification, the most popular is the \textit{dissimilarity representation} approach\cite{Pekal2005a}. It consists in using a set $R$ of $m$ reference instances, to build a $n \times m$ dissimilarity matrix. The elements of this matrix are the dissimilarities between the $n$ training instances in $T$ and the $m$ reference instances in $R$:
\begin{equation}
    \mathbf{D}(T,R) = \begin{bmatrix}
        d(\mathbf{x}_1, \mathbf{p}_1) & d(\mathbf{x}_1, \mathbf{p}_2) & \dots & d(\mathbf{x}_1, \mathbf{p}_m)\\
        d(\mathbf{x}_2, \mathbf{p}_1) & d(\mathbf{x}_2, \mathbf{p}_2) & \dots & d(\mathbf{x}_2, \mathbf{p}_m)\\
        \dots & \dots & \dots & \dots \\
        d(\mathbf{x}_n, \mathbf{p}_1) & d(\mathbf{x}_n, \mathbf{p}_2) & \dots & d(\mathbf{x}_n, \mathbf{p}_m)\\
     \end{bmatrix}
\end{equation}
where $d$ stands for a dissimilarity measure, $\mathbf{x}_i$ are the training instances and $\mathbf{p}_j$ are the reference instances. Even if $T$ and $R$ can be disjoint sets of instances, the most common is to take $R$ as a subset of $T$, or even as $T$ itself. In this work, for simplification purpose, and to avoid the selection of reference instances from $T$, we choose $R = T$. As a consequence the dissimilarity matrix $\mathbf{D}$ is always a symmetric $n \times n$ matrix.

Once such a squared dissimilarity matrix is built, there exist two main ways to use it for classification: the \textit{embedding} approach and the \textit{dissimilarity space} approach \cite{Pekal2005a}. The \textit{embedding} approach consists in embedding the dissimilarity matrix in a Euclidean vector space such that the distances between the objects in this space are equal to the given dissimilarities. Such an exact embedding is possible for every symmetric dissimilarity matrix with zeros on the diagonal \cite{Pekal2005a}. In practice, if the dissimilarity matrix can be transformed in a positive semi-definite (p.s.d.) similarity matrix, this can be done with kernel methods. This p.s.d. matrix is used as a pre-computed kernel, also called a kernel matrix. This method has been successfully applied with RFD along with SVM classifiers \cite{Gray2013, Cao2019a}. 

The second approach, the \textit{dissimilarity space} strategy, is more versatile and does not require the dissimilarity matrix to be transformed into a p.s.d. similarity matrix. It simply consists in using the dissimilarity matrix as a new training set. Indeed, each row $i$ of the matrix $\mathbf{D}$ can be seen as the projection of a training instance $\mathbf{x}_i$ into a dissimilarity space, where the $j^{th}$ dimension is the dissimilarity with the training instance $\mathbf{x}_j$. As a consequence, the matrix $\mathbf{D}(T,T)$ can be seen as the projection of the training set $T$ into this dissimilarity space, and can be fed afterward to any learning procedure. This method is much more straightforward than the \textit{embedding} approach as it can be used with any dissimilarity measurement, regardless its reflexivity or symmetry properties, and without transforming it into a p.s.d. similarity matrix.

In the following, the dissimilarity matrices built with the RFD measure are called RFD matrices and are noted $\mathbf{D}_H$ for short. It can be proven that the matrices derived from the initial RF proximity measure \cite{Breim2001, Verik2011} are p.s.d and can be used as pre-computed kernels in SVM classifiers \cite{Cao2019a}, following the \textit{embedding} approach. However, the proof does not apply if the matrices are obtained using the RFD measure \cite{Cao2019b}. This is the main reason we use the \textit{dissimilarity space} strategy in this work, as it allows more flexibility.

\subsection{Using dissimilarity spaces for multi-view learning}
\label{ssec:dissim_mvl}

In traditional supervised learning tasks, each instance is described by a single vector of $m$ features. For multi-view learning tasks, each instance is described by $Q$ different vectors. As a consequence, the task is to infer a model $h$:
\begin{equation}
	h : \mathcal{X}^{(1)} \times \mathcal{X}^{(2)} \times \dots \times \mathcal{X}^{(Q)} \rightarrow \mathcal{Y}
\end{equation}
where the $\mathcal{X}^{(q)}$ are the $Q$ input domains, i.e. the views. These views are generally of different dimensions, noted $m_1$ to $m_Q$. For such learning tasks, the training set $T$ is actually made up with $Q$ training subsets:
\begin{equation}
	T^{(q)} = \left\lbrace (\mathbf{x}^{(q)}_1,y_1), (\mathbf{x}^{(q)}_2,y_2),\dots,(\mathbf{x}^{(q)}_n,y_n) \right\rbrace, \forall q=1..Q
\end{equation}

The key principle of the proposed framework is to compute the RFD matrices $\mathbf{D}_H^{(q)}$ from each of the $Q$ training subsets $T^{(q)}$. For that purpose, each $T^{(q)}$ is fed to the RF learning procedure, resulting in $Q$ RF classifiers noted $H^{(q)}, \forall q=1..Q$. The RFD measure is then used to compute the $Q$ RFD matrices $\mathbf{D}_H^{(q)}, \forall q=1..Q$. 

Once these RFD matrices are built, they have to be merged in order to build the joint dissimilarity matrix $\mathbf{D}_H$ that will serve as a new training set for an additional learning phase. This additional learning phase can be realized with any learning algorithm, since the goal is to address the classification task. For simplicity and because they are as accurate as they are versatile, the same Random Forest method used to calculate the dissimilarities is also used in this final learning stage.

\begin{figure}
    \centering
    \includegraphics[width=0.9\textwidth]{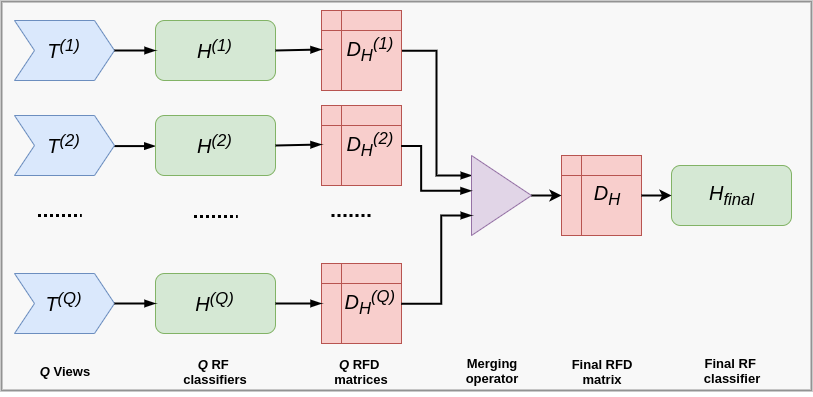}
    \caption{The RFD framework for multi-view learning.}
    \label{fig:train}
\end{figure}

Regarding the merging step, which is the main focus of the present work, it can be straightforwardly done by a simple average of the $Q$ RFD matrices:
\begin{equation}
    \mathbf{D}_H = \frac{1}{Q}\sum_{q=1}^{Q} \mathbf{D}_H^{(q)}
\end{equation}
The whole RFD based multi-view learning procedure is summarized in Algorithm \ref{algo:rfdis} and illustrated in Figure \ref{fig:train}.

\begin{algorithm}[ht]
    \small
    \SetAlgoLined
    \LinesNumbered
    \DontPrintSemicolon
    \KwIn{$T^{(q)}$, $\forall q=1..Q$: the $Q$ training sets, composed of $n$ instances}
    \KwIn{$RF(.)$: The Breiman's RF learning procedure}
    \KwIn{$RFD(.,.|.)$: the $RFD$ dissimilarity measure}
    \KwOut{$H^{(q)}$: $Q$ RF classifiers}
    \KwOut{$H_{final}$: the final RF classifier}
    \BlankLine
    \For{$q=1..Q$}{
        $H^{(q)} = RF(T^{(q)})$\;
        \tcp{Build the $n \times n$ RFD matrix $\mathbf{D}_H^{(q)}$:}
        \ForAll{$\mathbf{x}_i \in T^{(q)}$}{
            \ForAll{$\mathbf{x}_j \in T^{(q)}$}{
                $\mathbf{D}_H^{(q)}[i,j] = RFD(\mathbf{x}_i,\mathbf{x}_j | H^{(q)})$\;
            }
        }
    }
    \tcp{Build the $n \times n$ average RFD matrix $\mathbf{D}_H$:}
    $\mathbf{D}_H = \frac{1}{Q}\sum_{q=1}^{Q} \mathbf{D}_H^{(q)}$\;
    \tcp{Train the final classifier on $\mathbf{D}_H$:}
    $H_{final} = RF(\mathbf{D}_H)$\;
\caption{The RFD multi-view learning procedure\label{algo:rfdis}}
\end{algorithm}

As for the prediction phase, the procedure is very similar. For any new instance $\mathbf{x}$ to predict:
\begin{enumerate}
    \item Compute $d^{(q)}_H(\mathbf{x},\mathbf{x}_i), \forall \mathbf{x}_i \in T^{(q)}, \forall q=1..Q$, to form $Q$ $n$-sized dissimilarity vectors for $\mathbf{x}$. These vectors are the dissimilarity representations for $\mathbf{x}$, from each of the $Q$ views.
    \item Compute $d_H(\mathbf{x},\mathbf{x}_i) = \frac{1}{Q}\sum_{q=1}^{Q} d^{(q)}_H(\mathbf{x},\mathbf{x}_i), \forall i=1..n$, to form the $n$-sized vector that corresponds to the projection of $\mathbf{x}$ in the joint dissimilarity space.
    \item Predict the class of $\mathbf{x}$ with the classifier trained on $\mathbf{D}_H$.
\end{enumerate}

\section{Combining views with weighted combinations}
\label{sec:combi}

The average dissimilarity is a simple, yet meaningful way to merge the dissimilarity representations built from all the views. However, it intrinsically considers that all the views are equally relevant with regard to the task and that the resulting dissimilarities are as reliable as each other. This is likely to be wrong from our point of view. In multi-view learning problems, the different views are meant to be complementary in some ways, that is to say to convey different types of information regarding the classification task. These different types of information may not have the same contribution to the final predictions. That is the reason why it may be important to differentiate these contributions, for example with a weighted combination in which the weights would be defined according to the view reliability.

The calculation of these weights can be done following two paradigms: \textit{static weighting} and \textit{dynamic weighting}. The static weighting principle is to weight the views once for all, with the assumption that the importance of each view is the same for all the instances to predict. The dynamic weighting principle on the other way, aims at setting different weights for each instance to predict, with the assumption that the contribution of each view to the final prediction is likely to be different from one instance to another. 

\subsection{Static combination}
\label{ssec:sw}

Given a set of dissimilarity matrices $\{\mathbf{D}^{(1)},\mathbf{D}^{(2)}, \dots, \mathbf{D}^{(Q)}\}$ built from $Q$ different views, our goal is to find the best set of non-negative weights $\{w^{(1)},w^{(2)},$ $\dots, w^{(Q)}\}$, so that the joint dissimilarity matrix is:
\begin{equation}
    \label{jd}
    \mathbf{D} = \sum_{q=1}^{Q} w^{(q)}\mathbf{D}^{(q)}
\end{equation}
where $w^{(q)}\geq 0$ and $\sum_{q=1}^{Q} w^{(q)}=1$.\\


There exist several ways, proposed in the literature, to compute the weights of such a static combination of dissimilarity matrices. The most natural one is to deduce the weights from a quality score measured on each view. For example, this principle has been used for multi-scale image classification \cite{Li2012} where each view is a version of the image at a given scale, i.e. the weights are derived directly from the scale factor associated with the view. Obviously, this only makes sense with regard to the application, for which the scale factor gives an indication of the reliability for each view.

Another, more generic and classification-specific approach, is to evaluate the quality of the dissimilarity matrix using the performance of a classifier. This makes it possible to estimate whether a dissimilarity matrix sufficiently reflects class membership \cite{Duin2012,Li2012}. For example, one can train a SVM classifier from each dissimilarity matrix and use its accuracy as an estimation of the corresponding weights \cite{Li2012}. $k$NN classifiers are also very often used for that purpose \cite{Duin2012,Li2018}. The reason is that a good dissimilarity measure is expected to propose good neighborhoods, or in other words the most similar instances should belong to the same class.

Since kernel matrices can be viewed as similarity matrices, there are also few solutions in the literature of kernel methods that could be used to estimate the quality of a dissimilarity matrix. The most notable is the Kernel Alignment (KA) estimate \cite{Cristi2002} $A(\mathbf{K}_1, \mathbf{K}_2)$, for measuring the similarity between two kernel matrices $\mathbf{K}_1$ and $\mathbf{K}_2$:
\begin{equation}
    \label{46}
    A(\mathbf{K}_1, \mathbf{K}_2) = \frac{\langle \mathbf{K}_1, \mathbf{K}_2 \rangle_F }{\sqrt{\langle \mathbf{K}_1, \mathbf{K}_1 \rangle_F \langle \mathbf{K}_2, \mathbf{K}_2 \rangle_F}}
\end{equation}
where $K_i$ is a kernel matrix and where $\langle \cdot, \cdot \rangle_F$ is the Frobenius norm \cite{Cristi2002}.

In order to use the KA measure to estimate the quality of a given kernel matrix, a target matrix must be defined beforehand. This target matrix is an ideal theoretical similarity matrix, regarding the task. For example, for binary classification, the ideal target matrix is usually defined as $\mathbf{K}^* = \mathbf{y}\mathbf{y}^T$, where $\mathbf{y} = \{y_1, y_2, \dots, y_n\}$ are the true labels of the training instances, in $\{-1,+1\}$. Thus, each value in $\mathbf{K}^*$ is:
\begin{equation}\label{kl2a}
	\mathbf{K}^*_{ij} = 
        \begin{cases}
        1, & \text{if}\ y_i = y_j\\
        -1, & \text{otherwise}
        \end{cases}
\end{equation}
In other words, the ideal matrix is the similarity matrix in which instances are considered similar ($\mathbf{K}^*_{ij} = 1$) if and only if they belong to the same class. This estimate is transposed to multi-class classification problems as follows \cite{Camar2009}:
\begin{equation}\label{kl2}
	\mathbf{K}^*_{ij} = 
    	\begin{cases}
    	1, & \text{if}\ y_i = y_j\\
    	\frac{-1}{\mathcal{C}-1}, & \text{otherwise}
    	\end{cases}
\end{equation}
where $\mathcal{C}$ is the number of classes. \\

Both $k$NN and KA methods presented above are used in the experimental part for comparison purposes (cf. Section \ref{sec:xp}). However, in order to use the KA method for our problem, some adaptations are required. Firstly, the dissimilarity matrices need to be transformed into similarity matrices by $\mathbf{S}^{(q)} = 1-\mathbf{D}^{(q)}$. The following heuristic is then used to deduce the weight from the KA measure \cite{Qiu2009}:
\begin{equation}
    w^{(q)} = \frac{A(\mathbf{S}^{(q)},\mathbf{y}\mathbf{y}^T)}{\sum_{h=1}^Q A(\mathbf{S}^{(h)},\mathbf{y}\mathbf{y}^T)}
\end{equation}
Strictly speaking, for the similarity matrices $\mathbf{S}^{(q)}$ to be considered as kernel matrices, it must be proven that they are p.s.d. When such matrices are proven to be p.s.d, the KA estimates is necessarily non-negative, and the corresponding $w^{(q)}$ are also non-negative \cite{Cristi2002,Qiu2009}. However, as it is not proven that our matrices $\mathbf{S}^{(q)}$ built from $RFD$ are p.s.d., we propose to use the \textit{softmax} function to normalize the weights and to ensure they are strictly positive:
\begin{equation}
    w^{(q)} = \frac{exp(A(\mathbf{S}^{(q)},\mathbf{K}^*))}{\sum_{h=1}^Q exp(A(\mathbf{S}^{(h)},\mathbf{K}^*))}
\end{equation}
\newline


The main drawback of the methods mentioned above is that they evaluate the quality of the dissimilarity matrices based solely on the training set. This is the very essence of these methods, which are designed to evaluate (dis)similarity matrices built from a sample, e.g. the training set. However, this may cause overfitting issues when these dissimilarity matrices are used for classification purposes as it is the case in our framework. Ideally, the weights should be set from the quality of the dissimilarity representations estimated on an independent validation dataset. Obviously, this requires to have additional labeled instances. The method we propose in this section allows to estimate the quality of the dissimilarity representations without the use of additional validation instances. 

The idea behind our method is that the relevance of a RFD space is reflected by the accuracy of the RF classifier used to build it. This accuracy can be efficiently estimated with a mechanism called the Out-Of-Bag (OOB) error. This OOB error is an estimate supplied by the Bagging principle, known to be a reliable estimate of the generalization error \cite{Breim2001}. Since the RF classifiers in our framework are built with the Bagging principle, the OOB error can be used to estimate their generalization error without the need of an independent validation dataset.

Let us briefly explained here how the OOB error is obtained from a RF: let $B$ denote a Bootstrap sample formed by randomly drawing $p$ instances from $T$, with replacement. When $p=n$, $n$ being the number of instances in $T$, it can be proven that about one third of $T$, in average, will not be drawn to form $B$ \cite{Breim2001}. These instances are called the OOB instances of $B$. Using Bagging for growing a RF classifier, each tree in the forest is trained on a Bootstrap sample, that is to say using only about two thirds of the training instances. Similarly, each training instance $\mathbf{x}$ is used for growing about two thirds of the trees in the forest. The remaining trees are called the OOB trees of $\mathbf{x}$. The OOB error is the error rate measured on the whole training set by only using the OOB trees of each training instance.

Therefore, the method we propose to use consists in using the OOB error of the RF classifier trained on a view directly as its weight in the weighted combination. This method is noted $SW_{OOB}$ in the following.

\subsection{Dynamic combination}
\label{ssec:dyn}

In contrast to static weighting, dynamic weighting aims at assigning different weights to the views for each instance to predict \cite{Cruz2018}. The intuition behind using dynamic weighting in our framework is that the prediction for different instances may rely on different types of information, i.e. different views. In that case, it is crucial to use different weights for building the joint dissimilarity representation from one instance to predict to another.

However, such a dynamic weighting process is particularly complex in our framework. Let us recall that the framework we propose to use in this work is composed of two stages: (i) inferring the dissimilarity matrix from each view, and (ii) combining the per-view dissimilarity matrices to form a new training set. The weights we want to determine are the weights used to compute the final joint dissimilarity matrix in stage (ii). As a consequence, if these weights change for each instance to predict, the joint dissimilarity matrix must be completely recalculated and a new classifier must also be re-trained afterwards. This means that, for every new instance to predict, a whole training procedure has to be performed. This is computationally expensive and quite inefficient from our point of view.

To overcome this problem, we propose to use Dynamic Classifier Selection (DCS) instead of dynamic weighting. DCS is a generic strategy, amongst the  most successful ones in the Multiple Classifier Systems literature \cite{Cruz2018}. It typically aims at selecting one classifier in a pool of candidate classifiers, for each instance to predict. This is essentially done through two steps \cite{Britt2014}: (i) the generation of a pool of candidate classifiers and (ii) the selection of the most competent classifier in this pool for the instance to predict. The solutions we propose for these steps are illustrated in Figure \ref{fig:DCS}, the first step in the upper part and the second step in the lower part. The whole procedure is also detailed in Algorithm \ref{algo:dynrfd} and described in the following.

\subsubsection{Generation of the pool of classifiers}
\label{sssec:dcs_pool}

The generation of the pool is the first key step of DCS. As the aim is to select the most competent classifier on the fly for each given test instance, the classifiers in the pool must be as diverse and as individually accurate as possible. In our case, the challenge is not to create the diversity in the classifiers, since they are trained on different joint dissimilarity matrices, generated with different sets of weights. The challenge is rather to generate these different weight tuples used to compute the joint dissimilarity matrices. For such a task, a traditional grid search strategy could be used. However, the number of candidate solutions increases exponentially with respect to the number of views. For example, Suppose that we sample the weights with 10 values in $[0,1]$. For $Q$ views, it would result in $10^Q$ different weight tuples. Six views would thus imply to generate 1 million weight tuples and to train 1 million classifiers afterwards. Here again, this is obviously highly inefficient.\\

The alternative approach we propose is to select a subset of views for every candidate in the pool, instead of considering a weighted combination of all of them. By doing so, for each instance to predict, only the views that are considered informative enough are expected to be used for its prediction. The selected views are then combined by averaging. For example, if a problem is described with six views, there are $2^6-1=63$ possible combinations (the situation in which none of the views is selected is obviously ignored), which will result in a pool of 63 classifiers $\mathcal{H} = \{H_1, H_2, \dots, H_{63}\}$. Lines 1 to 6 of Algorithm \ref{algo:dynrfd} give a detailed implementation of this procedure.

\subsubsection{Evaluation and selection of the best classifier}

The selection of the most competent classifier is the second key step of DCS. Generally speaking, this selection is made through two steps \cite{Cruz2018}: (i) the definition of a region of competence for the instance to predict and (ii) the evaluation of each classifier in the pool for this region of competence, in order to select the most competent one.

The region of competence $\Theta_t$ of each instance $\mathbf{x}_t$ is the region used to estimate the competence of the classifiers for predicting that instance. The usual way to do so is to rely on clustering methods or to identify the $k$ nearest neighbors ($k$NN) of $\mathbf{x}_t$. For clustering \cite{Soare2006}, the principle is usually to define the region of competence as the closest cluster of $\mathbf{x}_t$, according to the distances of $\mathbf{x}_t$ to the centroids of the clusters. As the clusters are fixed once for all, many different instances might share the same region of competence. In contrast, $k$NN methods give different regions of competence from one instance to another, which allows for more flexibility but at the expense of a higher computational cost \cite{DeSou2008}.

The most important part of the selection process is to define the criterion to measure the competence level of each classifier in the pool. There are a lot of methods for doing so, that differ in the way they estimate the competence, using for example a ranking, the classifier accuracies, a data complexity measure, etc. \cite{Cruz2018}. Nevertheless, the general principle is most of the time the same: calculating the measure on the region of competence exclusively. We do not give an exhaustive survey of the way it can be done here, but briefly explain the most representative method, namely the Local Classifier Accuracy (LCA) method \cite{Woods1997}, as an illustration. 

The LCA method measures the local accuracy of a candidate classifier $H_i$, with respect to the prediction $\hat{y}_t$ of a given instance $\mathbf{x}_t$:
\begin{equation}\label{lca}
	w_{i,t} = \frac{\sum_{\mathbf{x}_k \in \Theta_{t,\hat{y}_t}} I( H_i(\mathbf{x}_k) = \hat{y}_t)}{\sum_{\mathbf{x}_k \in \Theta_t} I(y_k = \hat{y}_t)} 
\end{equation}
where $\Theta_t =\{\mathbf{x}_1, \dots \mathbf{x}_k, \dots, \mathbf{x}_K \}$ is the region of competence for $\mathbf{x}_t$, and $\Theta_{t,\hat{y}_t}$ is the set of instances from $\Theta_t$ that belong to the same class as $\hat{y}_t$. Therefore, $w_{i,t}$ represents the percentage of correct classifications within the region of competence, by only considering the instances for which the classifier predicts the same class as for $\mathbf{x}_t$. In this calculation, the instances in $\Theta_t$ generally come from a validation set, independent of the training set $T$ \cite{Cruz2018}. \\

The alternative method we propose here is to use a selection criterion that does not rely on an independent validation set, but rather relies on the OOB estimate. To do so, the region of competence is formed by the $k$ nearest neighbors of $\mathbf{x}_t$, amongst the training instances. These nearest neighbors are determined in the joint dissimilarity space with the RFD measure (instead of the traditional Euclidean distance). This is related to the fact that each candidate classifier is trained in this dissimilarity space, but also because the RFD measure is more robust to high dimensional spaces, contrary to traditional distance measures. Finally, the competence of each classifier is estimated with its OOB error on the $k$ nearest neighbors of $\mathbf{x}_t$. Lines 7 to 15 of Algorithm \ref{algo:dynrfd} give all the details of this process.

%



\begin{figure}
    \centering
    \includegraphics[width=0.9\textwidth]{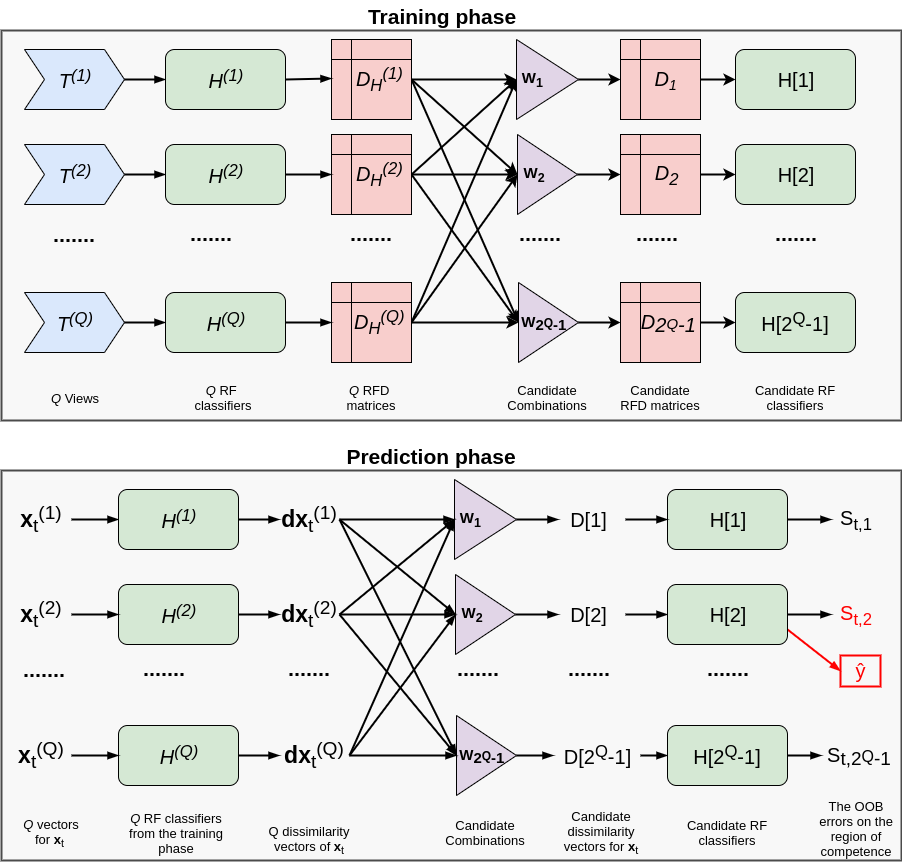}
    \caption{The $DCS_{RFD}$ procedure, with the training and prediction phases. The best candidate classifier that gives the final prediction for $\mathbf{x}_t$ is $H[2]$ in this illustration (in red).}
    \label{fig:DCS}
\end{figure}

To sum it up, the key mechanisms of the DCS method we proposed, noted $DCS_{RFD}$ and detailed in Algorithm \ref{algo:dynrfd}, are:
\begin{itemize}
    \item Create the pool of classifiers by using all the possible subsets of views, to avoid the expensive grid search for the weights generation (lines 4-5 of Algorithm \ref{algo:dynrfd}).
    \item Define the region of competence in the dissimilarity space by using the RFD dissimilarity measure, to circumvent the issues that arise from high dimensional spaces (lines 12-13 of Algorithm \ref{algo:dynrfd}).
    \item Evaluate the competence of each candidate classifier with its OOB error rate, so that no additional validation instances are required (line 14 of Algorithm \ref{algo:dynrfd}).
    \item Select the best classifier for $\mathbf{x}_t$ (lines 16-17 of Algorithm \ref{algo:dynrfd}).
\end{itemize}
These steps are also illustrated in Figure \ref{fig:DCS} with the generation of the pool of classifiers in the upper part, and with the evaluation and selection of the classifier in the lower part of the figure. For illustration purposes, the classifier ultimately selected for predicting the class of $\mathbf{x}_t$ is assumed to be the second candidate (in red).

\begin{algorithm}
    \small
    \SetAlgoLined
    \LinesNumbered
    \DontPrintSemicolon
    \KwIn{$T^{(q)}$, $\forall q=1..Q$: the $Q$ training sets, each composed of $n$ instances}
    \KwIn{$D^{(q)}$, $\forall q=1..Q$: $Q$ $n \times n$ RFD matrices, built from the $Q$ views}
    \KwIn{$H^{(q)}$, $\forall q=1..Q$: the $Q$ RF classifiers used to build the $D^{(q)}$}
    \KwIn{$RF(.)$: The RF learning procedure}
    \KwIn{$RFD(.,.|.)$: the $RFD$ measure}
    \KwIn{$k$: the number of neighbors to define the region of competence}
    \KwIn{$\mathbf{x}_t$: an instance to predict}
    \KwOut{$\hat{y}$: the prediction for $\mathbf{x}_t$}
    \BlankLine
    \tcp{1 - Generate the pool of classifiers:}
    $\{\mathbf{w}_0, \mathbf{w}_1, \dots, \mathbf{w}_{2^Q-1} \}=$ all the possible Q-sized 0/1 vectors\;
    $\mathcal{H}=$ an empty pool of classifiers\;
    \For{$i=1..2^Q-1$}{
        \tcp{The $i^{th}$ candidate classifier in the pool, $\mathbf{w}_{i}[q]$ being the $q^{th}$ value of $\mathbf{w}_{i}$, either equal to 1 or 0:}
        $D_i = \frac{1}{Q} \sum_{q=1}^Q D^{(q)} . \mathbf{w}_{i}[q]$\;
        $\mathcal{H}[i] = RF(D_i)$\;
    }
    \tcp{2 - Evaluate the candidate classifiers for $\mathbf{x}_t$}
    \For{$q=1..Q$}{
        \tcp{the $q^{th}$ dissimilarity representation of $\mathbf{x}_t$}
        $\mathbf{dx}_t^{(q)} = RFD(\mathbf{x}_t,\mathbf{x}_j|H^{(q)}), \forall \mathbf{x}_j \in T^{(q)}$ \;
    }
    $\mathcal{D}=$ an empty set of dissimilarity representations of $\mathbf{x}_t$ \;
    \For{$i=1..2^Q-1$}{
        \tcp{The averaged dissimilarity representation of $\mathbf{x}_t$:}
        $\mathcal{D}[i] = \frac{1}{Q} \sum_{q=1}^Q \mathbf{dx}_t^{(q)} . \mathbf{w}_{i}[q]$\;
        \tcp{The region of competence, $D_i[j,.]$ being the $j^{th}$ row of $D_i$ :}
        $\theta_{t,i} =$ the $k$NN according to $RFD(\mathcal{D}[i],D_i[j,.] | \mathcal{H}[i]), \forall j=1..n$\;
        \tcp{The competence of $\mathcal{H}(i)$ on $\theta_{t,i}$:}
        $S_{t,i} = OOB_{err}(\mathcal{H}[i],\theta_{t,i})$ \;
    }
    
    \tcp{3 - Select the best classifier for $\mathbf{x}_t$ and predict its class}
    $m = \argmax_i S_{t,i}$\;
    $\hat{y} = \mathcal{H}[m](\mathcal{D}[m])$\;
    \caption{The $DCS_{RFD}$ method\label{algo:dynrfd}}
\end{algorithm}

\section{Experiments}
\label{sec:xp}
\subsection{Experimental protocol}

Both the $SW_{OOB}$ and the $DCS_{RFD}$ methods are evaluated on several real-world multi-view datasets in the following, and compared to state-of-the-art methods: the simple average of the view-specific dissimilarity matrices as a baseline method and the two static weighting methods presented in Section \ref{ssec:sw}, namely the $3NN$ and the $KA$ methods.\\

The multi-view datasets used in this experiment are described in Table \ref{tab:data}. All these datasets are real-world multi-view datasets, supplied with several views of the same instances: \textit{NonIDH1}, \textit{IDHcodel}, \textit{LowGrade} and \textit{Progression} are medical imaging classification problems, with different families of features extracted from different types of radiographic images; \textit{LSVT} and \textit{Metabolomic} are two other medical related classification problems, the first one for Parkinson's disease recognition and the second one for colorectal cancer detection; \textit{BBC} and \textit{BBCSport} are text classification problems from news articles; \textit{Cal7}, \textit{Cal20}, \textit{Mfeat}, \textit{NUS-WIDE2}, \textit{NUS-WIDE3}, \textit{AWA8} and \textit{AWA15} are image classification problems made up with different families of features extracted from the images. More details about how these datasets have been constituted can be found in the paper (and references therein) cited in the caption of Table \ref{tab:data}.

\begin{table}[ht]
    \caption{Real-world multi-view datasets \cite{Cao2019a}. $^{\text a}$Imbalanced Ratio, i.e. the number of instances from the majority class over the number of instances from the minority class. 
    }
    {\begin{adjustbox}{max width=0.8\textwidth}
    \begin{tabular}{@{}lccccc@{}}
      \hline
      & features & instances & views & classes & IR$^{\text a}$ \\
      \hline
      AWA8
      & 10940 &640 &6 & 8& 1\\
      AWA15
      & 10940 &1200 & 6 & 15& 1\\
      BBC
      & 13628 &2012 &2 & 5& 1.34\\
      BBCSport
      & 6386 &544 & 2 & 5& 3.16\\
      Cal7
      & 3766 &1474 & 6 & 7& 25.74  \\
      Cal20
      & 3766 &2386 & 6 & 20& 24.18  \\
      IDHcodel
      & 6746 &67 & 5 & 2&2.94 \\
      LowGrade
      & 6746 &75 & 5 & 2&1.4 \\
      LSVT
      & 309 &126 & 4 & 2&2 \\
      Metabolomic
      & 476 &94 & 3 & 2& 1 \\
      Mfeat
      & 649 &600 & 6 & 10&1 \\
      NonIDH1
      & 6746 &84 & 5 & 2&3 \\
      NUS-WIDE2
      & 639 &442 & 5 & 2& 1.12 \\
      NUS-WIDE3
      & 639 &546 & 5 & 3&1.43 \\
      Progression
      & 6746 &84 & 5 & 2&1.68 \\
      \hline
    \end{tabular}
    \end{adjustbox}}
    \label{tab:data}
\end{table}

All the methods used in these experiments include the same first stage, i.e. building the RF classifiers from each view and building then the view-specific RFD matrices. Therefore, for a fair comparison on each dataset, all the methods use the exact same RF classifiers, made up with the same $512$ trees \cite{Cao2019a}. As for the other important parameters of the RF learning procedure, the $mtry$ parameter is set to $\sqrt{m_q}$, where $m_q$ is the dimension of the $q^{th}$ view, and all the trees are grown to their maximum depth (i.e. with no pre-pruning).

The methods compared in this experiment differ in the way they combine the view-specific RFD matrices afterwards. We recall below these differences:
\begin{itemize}
    \item $Avg$ denotes the baseline method for which the joint dissimilarity representation is formed by a simple average of the view-specific dissimilarity representations.
    \item $SW_{3NN}$ and $SW_{KA}$ both denote static weighting methods for determining $Q$ weights, one per view. The first one derives the weights from the performance of a $3NN$ classifier applied on each RFD matrix; the second one uses the KA method to estimate the relevancy of each RFD matrix in regards to the classification problem.
    \item $SW_{OOB}$ is the static weighting method we propose in this work and presented in Section \ref{ssec:sw}; it computes the weights of each view from the OOB error rate of its RF classifier.
    \item $DCS_{RFD}$ is the dynamic selection method we propose in this work and presented in Section \ref{ssec:dyn}; it computes different combinations of the RFD matrices for each instance to predict based on its $k$ nearest neighbors, with $k=7$ following the recommendation in the literature \cite{Cruz2018}.
\end{itemize}
After each method determine a set of $Q$ weights, the joint RFD matrix is computed. This matrix is then used as a new training set for a RF classifier learnt with the same parameters as above (512 trees, $mtry = \sqrt{n}$ with $n$ the number of training instances, fully grown trees).

As for the pre-processing of the datasets, a stratified random splitting procedure is repeated 10 times, with 50\% of the instances for training and 50\% for testing. The mean accuracy, with standard deviations, are computed over the 10 runs and reported in Table \ref{tab:ch52}. Bold values in this table are the best average performance obtained on each dataset.

\begin{table}[ht]
    \caption{Accuracy (mean $\pm$ standard deviation) and average ranks}
    {\begin{adjustbox}{max width=\textwidth}
    \begin{tabular}{@{}lccccc@{}} 
    	\hline
        & Avg &$SW_{3NN}$  & $SW_{KA}$ & $SW_{OOB}$ & $DCS_{RFD}$ \\ 
        \hline
        AWA8 &
         $56.22\% \pm 1.01$ &
         $56.22\% \pm 0.99$ &
         $56.12\% \pm 1.42$ &
         $56.59\% \pm 1.41$ &
         $\mathbf{57.28\% \pm 1.49}$
         \vspace*{0.0mm} \\
        AWA15 &
         $38.23\% \pm 0.83$ &
         $38.13\% \pm 0.87$ &
         $38.27\% \pm 1.05$ &
         $38.23\% \pm 1.26$ &
         $\mathbf{38.82\% \pm 1.56}$
         \vspace*{0.0mm} \\
        BBC &
         $95.46\% \pm 0.65$ &
         $\mathbf{95.52\% \pm 0.64}$ &
         $95.36\% \pm 0.74$ &
         $95.46\% \pm 0.60$ &
         $95.42\% \pm 0.59$
         \vspace*{0.0mm} \\
        BBCSport &
         $90.18\% \pm 1.96$ &
         $90.29\% \pm 1.83$ &
         $90.26\% \pm 1.78$ &
         $90.26\% \pm 1.95$ &
         $\mathbf{90.44\% \pm 1.89}$
         \vspace*{0.0mm} \\
        Cal7 &
         $96.03\% \pm 0.53$ &
         $96.10\% \pm 0.57$ &
         $\mathbf{96.11\% \pm 0.60}$ &
         $96.10\% \pm 0.60$ &
         $94.65\% \pm 1.09$
         \vspace*{0.0mm} \\
        Cal20 &
         $89.76\% \pm 0.80$ &
         $89.88\% \pm 0.82$ &
         $89.77\% \pm 0.68$ &
         $\mathbf{90.00\% \pm 0.71}$ &
         $89.15\% \pm 0.97$
         \vspace*{0.0mm} \\
        IDHCodel &
         $76.76\% \pm 3.59$ &
         $77.06\% \pm 3.43$ &
         $77.35\% \pm 3.24$ &
         $76.76\% \pm 3.82$ &
         $\mathbf{77.65\% \pm 3.77}$
         \vspace*{0.0mm} \\
        LowGrade &
         $63.95\% \pm 5.62$ &
         $62.56\% \pm 6.10$ &
         $63.95\% \pm 3.57$ &
         $63.95\% \pm 5.01$ &
         $\mathbf{65.81\% \pm 5.31}$
         \vspace*{0.0mm} \\
        LSVT &
         $84.29\% \pm 3.51$ &
         $84.29\% \pm 3.65$ &
         $84.60\% \pm 3.54$ &
         $\mathbf{84.76\% \pm 3.63}$ &
         $84.44\% \pm 3.87$
         \vspace*{0.0mm} \\
        Metabolomic &
         $69.17\% \pm 5.80$ &
         $68.54\% \pm 5.85$ &
         $70.00\% \pm 4.86$ &
         $70.00\% \pm 6.12$ &
         $\mathbf{70.21\% \pm 4.85}$
         \vspace*{0.0mm} \\
        Mfeat &
         $97.53\% \pm 1.00$ &
         $97.53\% \pm 1.09$ &
         $97.53\% \pm 1.09$ &
         $97.57\% \pm 1.01$ &
         $\mathbf{97.63\% \pm 0.99}$
         \vspace*{0.0mm} \\
        NonIDH1 &
         $80.70\% \pm 3.76$ &
         $80.47\% \pm 3.32$ &
         $80.00\% \pm 3.15$ &
         $\mathbf{80.93\% \pm 4.00}$ &
         $79.77\% \pm 2.76$
         \vspace*{0.0mm} \\
        NUS-WIDE2 &
         $92.82\% \pm 1.93$ &
         $92.86\% \pm 1.88$ &
         $92.60\% \pm 2.12$ &
         $92.97\% \pm 1.72$ &
         $\mathbf{93.30\% \pm 1.58}$
         \vspace*{0.0mm} \\
        NUS-WIDE3 &
         $80.32\% \pm 1.95$ &
         $79.95\% \pm 2.40$ &
         $80.09\% \pm 2.07$ &
         $80.14\% \pm 2.20$ &
         $\mathbf{80.77\% \pm 2.06}$
         \vspace*{0.0mm} \\
        Progression &
         $65.79\% \pm 4.71$ &
         $65.79\% \pm 4.71$ &
         $65.79\% \pm 4.99$ &
         $66.32\% \pm 4.37$ &
         $\mathbf{66.84\% \pm 5.29}$
         \vspace*{0.0mm} \\
         \hline 
        Avg rank
        &3.67
        &3.50
        &3.30
        &2.40
        &2.13 \\ 
        \hline
    \end{tabular}
    \end{adjustbox}}
    \label{tab:ch52}
\end{table}

\subsection{Results and discussion}

\begin{figure}
    \centerline{\includegraphics[width=0.65\textwidth]{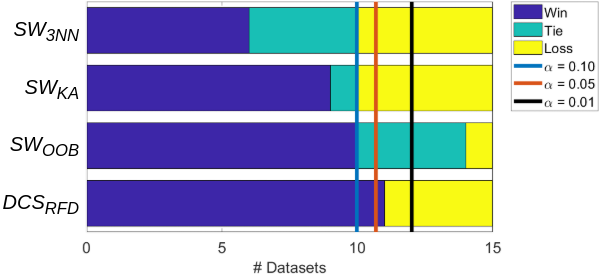}}
    \caption{Pairwise comparison between each method and the baseline $Avg$. The vertical lines are the level of statistical significance according to the Sign test.}
    \label{4stat1}
\end{figure}

The first observation one can make from the results gathered in Table \ref{tab:ch52} is that the best performance are obtained with one of the two proposed methods for 13 over the 15 datasets. This is confirmed by the average ranks that place these two methods in the first two positions. To better assess the extent to which these differences are significant, a pairwise analysis based on the Sign test is computed on the number of wins, ties and losses between the baseline method $Avg$ and all the other methods. The result is shown in Figure \ref{4stat1}.

From this statistical test, one can observe that none of the static weighting methods allows to reach the significance level of wins over the baseline method. It indicates that the simple average combination, when using dissimilarity representations for multi-view learning, is a quite strong baseline. It also underlines that all views are globally relevant for the final classification task. There is no view that is always irrelevant, for all the predictions. 

Figure \ref{4stat1} shows also that the dynamic selection method proposed in this work is the only method that predominantly improves the accuracy over this baseline, till reaching the level of statistical significance. From our point of view, it shows that all the views do not participate in the same extent to the good prediction of every instance. Some instances are better recognized when the dissimilarities are computed by relying on some views more than on the others. These views are certainly not the same ones from one instance to another, and some instances may need the dissimilarity information from all the views at some point. Nevertheless, this highlights that the confusion between the classes is not always consistent from one view to another. In that sense, the views complement each others, and this can be efficiently exploited for multi-view learning provided that we can identify the views that are the most reliable for every instance, one by one.

\section{Conclusion}
Multi-view data are now very common in real world applications. Whether they arise from multiple sources or from multiple feature extractors, the different views are supposed to provide a more accurate and complete description of objects than a single description would do. Our proposal in this work was to address multi-view classification tasks using dissimilarity strategies, which give an efficient way to handle the heterogeneity of the multiple views.

The general framework we proposed consists in building an intermediate dissimilarity representation for each view, and in combining these representations afterwards for learning. The key mechanism is to use Random Forest classifiers to measure the dissimilarities. Random Forests embed a (dis)similarity measure that takes the class membership into account in such a way that instances from the same class are similar. The resulting dissimilarity representations can be efficiently merged since they are fully comparable from one view to another.  

Using this framework, our main contribution was to propose a dynamic view selection method that provides a better way of merging the per-view dissimilarity representations: a subset of views is selected for each instance to predict, in order to take the decision on the most relevant views while at the same time ignoring as much as possible the irrelevant views. This subset of views is potentially different from one instance to another, because all the views do not contribute at the same extent to the prediction of each instance. This has been confirmed on several real-world multi-view datasets, for which the dynamic combination of views has allowed to obtain much better results than static combination methods. 

However, in its current form, the dynamic selection method proposed in this chapter strongly depends on the number of candidate classifiers in the pool. To allow for more versatility, it could be interesting to decompose each view into several sub-views. This could be done for example, by using Bagging and Random Subspaces principles before computing the view-specific dissimilarities. In such a way, the dynamic combination could only select some specific part of each view, instead of considering the views as a whole.

\section*{Acknowledgement}
\noindent This work is part of the DAISI project, co-financed by the European Union with the European Regional Development Fund (ERDF) and by the Normandy Region.

\bibliographystyle{abbrv}
\bibliography{main}

\begin{thebibliography}{10}

\bibitem{Biau2016}
G.~Biau and E.~Scornet.
\newblock A random forest guided tour.
\newblock {\em TEST}, 25:197--227, 2016.

\bibitem{Breim2001}
L.~Breiman.
\newblock Random forests.
\newblock {\em Machine Learning}, 45(1):5--32, 2001.

\bibitem{Britt2014}
A.~S. Britto~Jr, R.~Sabourin, and L.~E. Oliveira.
\newblock Dynamic selection of classifiers - a comprehensive review.
\newblock {\em Pattern Recognition}, 47(11):3665--3680, 2014.

\bibitem{Camar2009}
J.~E. Camargo and F.~A. Gonz\'{a}lez.
\newblock A multi-class kernel alignment method for image collection
  summarization.
\newblock In {\em Proceedings of the 14th Iberoamerican Conference on Pattern
  Recognition: Progress in Pattern Recognition, Image Analysis, Computer
  Vision, and Applications (CIARP)}, pages 545--552. Springer-Verlag, 2009.

\bibitem{Cao2019b}
H.~Cao.
\newblock {\em Random Forest For Dissimilarity Based Multi-View Learning:
  Application To Radiomics}.
\newblock PhD thesis, University of Rouen Normandy, 2019.

\bibitem{Cao2019a}
H.~Cao, S.~Bernard, R.~Sabourin, and L.~Heutte.
\newblock Random forest dissimilarity based multi-view learning for radiomics
  application.
\newblock {\em Pattern Recognition}, 88:185--197, 2019.

\bibitem{Chen2017}
X.~Chen, H.~Ma, J.~Wan, B.~Li, and T.~Xia.
\newblock Multi-view 3d object detection network for autonomous driving.
\newblock In {\em IEEE Conference on Computer Vision and Pattern Recognition
  (CVPR)}, pages 6526--6534, 2017.

\bibitem{Costa2019}
Y.~M.~G. Costa, D.~Bertolini, A.~S. Britto, G.~D.~C. Cavalcanti, and L.~E.~S.
  de~Oliveira.
\newblock The dissimilarity approach: a review.
\newblock {\em Artificial Intelligence Review}, pages 1--26, 2019.

\bibitem{Cristi2002}
N.~Cristianini, J.~Shawe-Taylor, A.~Elisseeff, and J.~S. Kandola.
\newblock On kernel-target alignment.
\newblock In {\em Advances in Neural Information Processing Systems (NeurIPS)},
  pages 367--373, 2002.

\bibitem{Cruz2018}
R.~M. Cruz, R.~Sabourin, and G.~D. Cavalcanti.
\newblock Dynamic classifier selection: Recent advances and perspectives.
\newblock {\em Information Fusion}, 41:195--216, 2018.

\bibitem{DeSou2008}
M.~C. De~Souto, R.~G. Soares, A.~Santana, and A.~M. Canuto.
\newblock Empirical comparison of dynamic classifier selection methods based on
  diversity and accuracy for building ensembles.
\newblock In {\em IEEE International Joint Conference on Neural Networks
  (IJCNN)}, pages 1480--1487. IEEE, 2008.

\bibitem{Duin2012}
R.~P. Duin and E.~Pekalska.
\newblock The dissimilarity space: Bridging structural and statistical pattern
  recognition.
\newblock {\em Pattern Recognition Letters}, 33(7):826--832, 2012.

\bibitem{Englu2012}
C.~Englund and A.~Verikas.
\newblock A novel approach to estimate proximity in a random forest: An
  exploratory study.
\newblock {\em Expert Systems with Applications}, 39(17):13046--13050, 2012.

\bibitem{Delga2014}
M.~Fern\'{a}ndez-Delgado, E.~Cernadas, S.~Barro, and D.~Amorim.
\newblock Do we need hundreds of classifiers to solve real world classification
  problems?
\newblock {\em Journal of Machine Learning Research}, 15:3133--3181, 2014.

\bibitem{Gray2013}
K.~R. Gray, P.~Aljabar, R.~A. Heckemann, A.~Hammers, and D.~Rueckert.
\newblock Random forest-based similarity measures for multi-modal
  classification of alzheimer's disease.
\newblock {\em NeuroImage}, 65:167--175, 2013.

\bibitem{Li2018}
D.~Li and Y.~Tian.
\newblock Survey and experimental study on metric learning methods.
\newblock {\em Neural Networks}, 2018.

\bibitem{Li2012}
Y.~Li, R.~P. Duin, and M.~Loog.
\newblock Combining multi-scale dissimilarities for image classification.
\newblock In {\em International Conference on Pattern Recognition (ICPR)},
  pages 1639--1642. IEEE, 2012.

\bibitem{Pekal2005a}
E.~Pekalska and R.~P.~W. Duin.
\newblock {\em The Dissimilarity Representation for Pattern Recognition:
  Foundations And Applications (Machine Perception and Artificial
  Intelligence)}.
\newblock World Scientific Publishing Co., Inc., 2005.

\bibitem{Qiu2009}
S.~Qiu and T.~Lane.
\newblock A framework for multiple kernel support vector regression and its
  applications to sirna efficacy prediction.
\newblock {\em IEEE/ACM Transactions on Computational Biology and
  Bioinformatics (TCBB)}, 6(2):190--199, 2009.

\bibitem{Rokac2016}
L.~Rokach.
\newblock Decision forest: Twenty years of research.
\newblock {\em Information Fusion}, 27:111--125, 2016.

\bibitem{Smith2014}
M.~R. Smith, T.~Martinez, and C.~Giraud-Carrier.
\newblock An instance level analysis of data complexity.
\newblock {\em Machine Learning}, 95(2):225--256, 2014.

\bibitem{Soare2006}
R.~G. Soares, A.~Santana, A.~M. Canuto, and M.~C.~P. de~Souto.
\newblock Using accuracy and diversity to select classifiers to build
  ensembles.
\newblock In {\em IEEE International Joint Conference on Neural Network
  (IJCNN)}, pages 1310--1316. IEEE, 2006.

\bibitem{Verik2011}
A.~Verikas, A.~Gelzinis, and M.~Bacauskiene.
\newblock Mining data with random forests: A survey and results of new tests.
\newblock {\em Pattern Recognition}, 44(2):330 -- 349, 2011.

\bibitem{Woods1997}
K.~Woods, W.~P. Kegelmeyer, and K.~Bowyer.
\newblock Combination of multiple classifiers using local accuracy estimates.
\newblock {\em IEEE transactions on pattern analysis and machine intelligence},
  19(4):405--410, 1997.

\end{thebibliography}

\end{document}